\documentclass[conference]{IEEEtran}
\IEEEoverridecommandlockouts
\usepackage{cite}
\usepackage{amsmath,amssymb,amsfonts}
\usepackage{algorithmic}
\usepackage{graphicx}
\usepackage{textcomp}
\usepackage{url}
\usepackage{xcolor}
\def\BibTeX{{\rm B\kern-.05em{\sc i\kern-.025em b}\kern-.08em
    T\kern-.1667em\lower.7ex\hbox{E}\kern-.125emX}}

\newcommand{\linebreakand}{%
  \end{@IEEEauthorhalign}
  \hfill\mbox{}\par
  \mbox{}\hfill\begin{@IEEEauthorhalign}
}
\begin{document}

\title{%VolTA-3D - Volumetric Token Alignment for Diverse Learning from 3D Brian MRIs for Downstream Classification and Segmentation, using a Vision Transformer Architecture
VolTA-3D: Self-Supervised Learning for Brain MRI using 3D Volumetric Token Alignment %for Downstream Classification and Segmentation, using a Vision Transformer Architecture\\
%{\footnotesize }
%\thanks{}
}

\author{\IEEEauthorblockN{Amy Makawana\textsuperscript{1}, Abhijeet Parida\textsuperscript{2}, Marius George Linguraru\textsuperscript{2,3},}
\IEEEauthorblockN{Julia Ive\textsuperscript{1}, Syed Muhammad Anwar\textsuperscript{2,3*}}

% \and
% \IEEEauthorblockN{Abhijeet Parida}
% \IEEEauthorblockA{\textit{Sheikh Zayed Institute,}\\
% \textit{Children’s National Hospital}\\
% Washington, DC, USA 
% %pabhijeet@childrensnational.org
% }
% \and
% \IEEEauthorblockN{Marius G. Linguraru}
% \IEEEauthorblockA{\textit{Sheikh Zayed Institute,} \\
% \textit{Children’s National Hospital}\\
% Washington, DC, USA 
% %pabhijeet@childrensnational.org
% }
% \linebreakand
% \IEEEauthorblockN{Julia Ive}
% \IEEEauthorblockA{\textit{Institute of Health Informatics} \\
% \textit{University College London,}
% London, UK \\
% %j.ive@ucl.ac.uk
% }
% \and
% %\IEEEauthorblockN{Syed M. Anwar}
\\\textsuperscript{1}{Institute of Health Informatics, University College London, London, UK}
\\\textsuperscript{2}{Sheikh Zayed Institute for Pediatric Surgical Innovation, Children’s National Hospital, Washington, DC USA}
\\\textsuperscript{3}{School of Medicine and Health Sciences, George Washington University, Washington, DC, USA}
\\
{\footnotesize\textsuperscript{*}sanwar@childrensnational.org}

}

\maketitle

\begin{abstract}

 Self-supervised learning (SSL) has advanced medical image analysis by enabling representation learning from large unlabelled data. However, in brain magnetic resonance imaging (MRI), most 3D models remain specialised for either segmentation or classification, limiting their ability to generalise across datasets, imaging protocols, and downstream tasks. This lack of transferability constrains the clinical utility of 3D MRI models, despite the availability of unlabelled volumetric data. We present \textit{VolTA-3D}, a self-supervised 3D Vision Transformer framework designed to learn transferable volumetric representations. VolTA-3D jointly aligns global class-style tokens and local patch tokens within a student–teacher paradigm and enforces fine-grained structural reconstruction. This combined global–local alignment addresses the limited semantic diversity and subtle anatomical variation characteristic of brain MRI, which challenge existing SSL approaches. We evaluate VolTA-3D on multiple out-of-distribution downstream tasks, including hippocampal segmentation and classification of sex and Alzheimer’s disease versus healthy controls. Across all tasks, representations learned by VolTA-3D outperform randomly initialised baselines, demonstrating improved transferability and robustness under domain shift. Hence jointly enforcing global semantic consistency and local structural learning during pretraining enables broader concept learning from unlabelled brain MRI data. Overall VolTA-3D supports effective multi-task downstream performance without task-specific pretraining, a step towards generalisable and clinically viable 3D models.
\end{abstract}


\section{Introduction}
Deep learning is increasingly applied to brain magnetic resonance imaging (MRI) for disease classification and structural segmentation, where manual interpretation remains slow and variable~\cite{Kwee2021}. These tasks are essential for observing anatomical changes to understand disease progression, planning surgical intervention, and delineating structures and regions, among other uses~\cite{Ivana2015}. However, most existing brain MRI models are highly task-specific, limiting their ability to generalise across datasets, imaging protocols, and downstream tasks—an essential requirement for reliable clinical deployment~\cite{1}.

A central factor limiting generalisability is the quality of learned representations. Brain anatomy and pathology are inherently 3D (three-dimensional), and models that fail to exploit volumetric structure risk learning acquisition-specific or slice-dependent features that do not transfer well across domains~\cite{9557808}. While 2D (two-dimensional) models are more widely adopted due to lower computational cost, they lose critical inter-slice context and global anatomical structure, which are more stable across scanners and protocols. As a result, 3D models consistently outperform 2D approaches in capturing anatomically meaningful features and improving cross-domain robustness, even in reduced data settings~\cite{Avesta2023}. 

Despite their representational advantages, supervised training of 3D brain MRI models is constrained by the lack of high-quality and diverse annotated medical imaging data, and where more diverse data is available, model training architectures are not always built to leverage these effectively~\cite{Wang2023}. Manual annotation of volumetric MRI is time-consuming and expensive, resulting in a far greater availability of unlabelled MRI data than reliably annotated datasets ~\cite{Ivana2015}. 

3D models additionally have the advantage of greater global context learning which can be further enhanced using vision transformers (ViTs)~\cite{Krishnan2024}. ViT models excel in capturing global structure, outperforming the more traditionally used convolutional neural networks (CNNs), but they often require substantial data for training, and their application to brain MRI is limited~\cite{KHALIQ2025108586}. 

Self-supervised learning (SSL) offers a promising solution to the lack of labelled medical data by enabling representation learning from the more abundant unlabelled MRI data collections~\cite{1}. However, SSL for brain MRI remains challenging due to the limited semantic diversity of brain anatomy and the subtle nature of clinically relevant variations~\cite{Chinn2021.05.22.21257645}. This motivates development of novel 3D SSL training strategies that can learn transferable volumetric representations from heterogeneous datasets, where scanner and protocol variation hinder generalisation to out of domain samples. 

We propose the novel Vit-based VolTA-3D (volumetric token alignment for 3D imaging) pretraining methodology to leverage diverse concept learning from 3D MRI sourced from various centres and scanners. Utilising both global and local token alignment across varying views of an MRI, alongside Gaussian-mixture masked-modelling (GMML), we show enhanced performance compared to randomly initialised baselines across both segmentation and classification tasks. Our results demonstrate the potential of this methodology to adapt to diverse downstream tasks, paving the way for larger-scale pretraining.

\noindent Our main contributions are:
\begin{itemize}
    \item We introduce \textit{VolTA-3D}, the first 3D extension of the Diverse Concept Modelling (DiCoM) framework~\cite{dicom2024}, enabling joint class-style token and patch-token alignment, with group masked model learning, for volumetric MRI. This integrates global semantic alignment with local structural reconstruction within a unified 3D ViT student–teacher paradigm.
    \item We demonstrate that VolTA-3D learns transferable 3D MRI representations, outperforming randomly initialised 3D ViT, Swin Transformer, and CNN baselines on out-of-distribution segmentation and classification tasks, demonstrating robustness to domain shift; a key challenge in clinical deployment. 
    \item We further demonstrate VolTA-3D outperforms baselines when trained on reduced data subsets, indicating utility in limited-data settings. 
\end{itemize}

\begin{figure*}[!ht]
    \centering
    \includegraphics[width=0.95\linewidth]{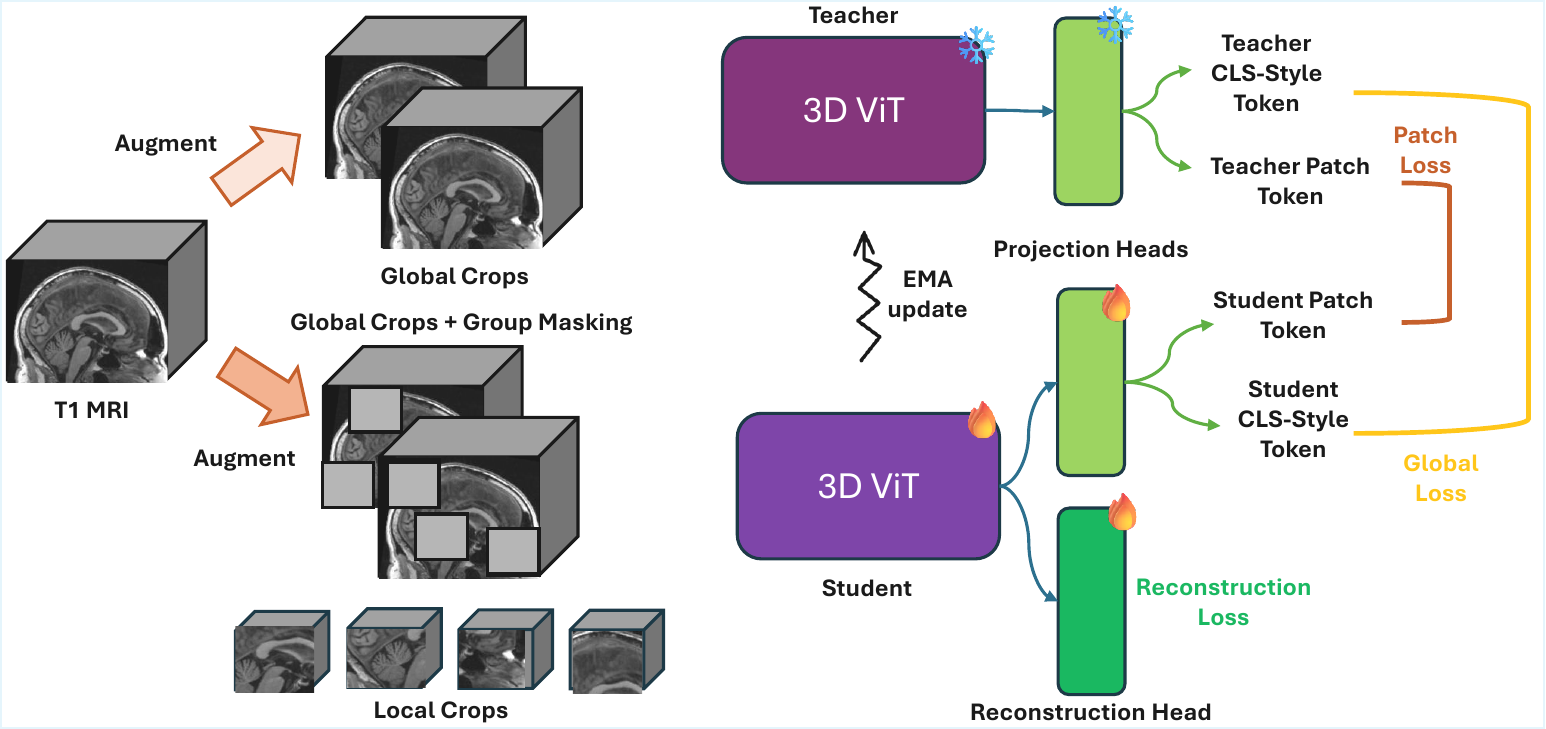}
    \caption{ VolTA-3D pretraining pipeline. A T1 MRI is geometrically augmented to produce two global crops with mild intensity for the teacher and stronger intensity with masking for the student—which also receives clean local crops. Both networks share the wrapped ViTAutoEnc encoder with a projection head, and the student additionally performs reconstruction.  }
    \label{fig:architecture}
\end{figure*}

\section{Methods}
\subsection{Data and Preprocessing}
To pretrain VolTA-3D, we used 2,002 diverse 3D T1/T1n MRIs: 773 from the Alzheimer’s Disease Neuro-imaging Initiative (ADNI) dataset~\cite{ADNI} spanning cognitively normal, significant memory concern, early mild cognitive impairment, late mild cognitive impairment, and Alzheimer’s disease (AD) patient groups, and 1,229 from the Brain Tumor Segmentation (BraTS) dataset~\cite{deverdier20242024braintumorsegmentation} from adult glioma, adult metastasis and peadiatric glioma. 

For downstream tasks, HarP-labelled ADNI scans (135 MRIs)~\cite{boccardi2015harp} were used for hippocampus segmentation (all patients excluded from ADNI pretraining). Sex classification used 619 healthy T1 MRIs (342 male and 277 female) from IXI\footnote{\url{https://brain-development.org/ixi-dataset/}}, and AD classification used 420 T1 MRIs (347 healthy and 73 AD) from the AIBL dataset~\cite{ellis2009aibl}. To match BraTS preprocessing~~\cite{kofler2020brats}, ADNI 2D DICOMs (Digital Imaging and Communications in Medicine files) were converted to NIfTI (Neuroimaging Informatics Technology Initiative  files), re-oriented to right-anterior-superior orientation, resampled to  1mm$^3$ isotropic resolution. All MRI were resized/padded to  $144\times192\times192$ with intensities scaled to  $[0, 1]$.

\subsection{VolTA-3D Framework and pretraining}

In general, SSL aims to extract information from data without ground truth labels to minimise human bias. We extend the SSL DiCoM [6] framework into 3D for VolTA-3D, incorporating: 1) GMML, 2) local and global alignment, and 3) knowledge distillation. The high-level VoLTA-3D pipeline is presented in Fig.~\ref{fig:architecture}. VolTA-3D is built around MONAI’s ViTAutoEnc, a 3D Vision Transformer autoencoder designed for volumetric inputs. The model partitions each MRI volume into 3D patches, embeds them as a sequence of tokens, and processes these tokens with Transformer encoder blocks to capture contextual relationships across the volume. We use the encoder output as the shared representation for both the student and teacher networks, whilst retaining the student's decoder to support reconstruction from corrupted inputs. As our implementation outputs patch tokens rather than a native CLS token, we add a projection head containing a learned CLS-style summariser, and lightweight transformer aggregator, to produce a global representation alongside the patch-level features.
 
\noindent \textbf{GMML:} We used the ViT encoder and integrated decoder for the student network ($s$). Using GMML, we randomly replace 50\% of the original MRI, $x$, with zeros/noise.  The masked MRIs are passed through the encoder and decoder, and the $\ell2$ loss  is computed over the masked voxels to reconstruct the original MRI, $\bar{x}$.
\begin{equation}
\mathcal{L}_{rec} = \ell2(x, \bar{x}).
\end{equation}

\noindent \textbf{CLS-Style Token:} As our implementation of ViTAutoEnc returns only patch tokens natively, we built a CLS-style summariser to enable global token alignment. Rather than simply averaging the patch tokens to create a summary token, we use a learned token which is updated through self-attention. The summariser comprises three main parts: a learned CLS-style token, a lightweight transformer aggregation block, and a projection branch. The CLS-style token is a trainable vector with shape [1, 1, 768], expanded to [B, 1, 768] for each batch, and prepended to the patch token sequence. This sequence is passed through the transformer aggregation block which contains layer normalisation, multi-head self attention (8 heads), a residual connection, a second layer normalisation, a multilayer perceptron (MLP) with Gaussian Error Linear Unit (GELU) activation, and a second residual connection. This allows the CLS-style token to attend to all patch tokens and gain global information. The projection branch takes the CLS-style token as the global summary, passes it through an MLP, undergoes L2-normalisation, and is then projected (using a weight normalised linear layer) to produce the final global representation used for CLS-level alignment.

\noindent \textbf{Local and Global Alignment:} To encourage representation consistency across multiple views of the same MRI, we enforce two alignment objectives: (1) \emph{global alignment} between the high-level \texttt{CLS} style token produced by the teacher and student ViTs, and (2) \emph{local alignment} between the ViT's per-token projections on the global crops. These two terms correspond to the classification consistency loss $\mathcal{L}_{\text{global}}$ and local alignment loss $\mathcal{L}_{\text{local}}$, respectively.

Given a set of $V$ augmented 3D views (2 global + local) of the same MRI, the teacher provides probability distributions $\{t^{(i)}\}_{i=1}^{2}$ , while the student produces logits $\{s^{(j)}\}_{j=1}^{V}$.  
Using self-distillation, each teacher global view is aligned with all student views \emph{except} itself. For a teacher distribution $t^{(i)}$ and a student logit vector $s^{(j)}$,  where $k$ indexes the dimensions of the output distribution, the consistency is computed using a cross-entropy objective:
\begin{equation}
\ell_{\text{cls}}^{(i,j)} = 
- \sum_{k} t^{(i)}_k \log \left( \text{softmax}(s^{(j)})_k \right),
\quad j \neq i.
\end{equation}

Here, $t^{(i)}$ denotes the teacher probability distribution obtained by centering the raw logits with an EMA centre vector $c$ and sharpening with teacher temperature $\tau_t$, following \cite{DBLP:journals/corr/abs-2104-14294}. The student softmax is similarly sharpened with a fixed temperature $\tau_s = 0.1$.

The global alignment loss $\mathcal{L}_{\text{global}}$ is therefore the average over all valid teacher--student view pairs:
\begin{equation}
\mathcal{L}_{\text{global}} =
\frac{1}{2(V-1)}
\sum_{i=1}^{2} \sum_{\substack{j=1 \\ j \neq i}}^{V}
\ell_{\text{cls}}^{(i,j)} .
\end{equation}
$\mathcal{L}_{\text{global}}$ encourages multi-view semantic consistency of multiple views of the MRI. 

Beyond global alignment, the student and teacher also produce patch-level local embeddings, denoted $s_{\text{patch}}$ and $t_{\text{patch}}$. To improve training stability, we sharpen the teacher and student outputs and apply centering to the teacher outputs: 
\begin{equation}
\tilde{t}_{\text{patch}} 
= 
\text{softmax}\!\left( 
\frac{t_{\text{patch}} - c}{\tau_t} 
\right),\  \tilde{s}_{\text{patch}} = \frac{s_{\text{patch}}}{\tau_s},
\end{equation}
where $c$ is a running mean, $\tau_s$ and $\tau_t$ are temperature scaling coefficients.  

The patch-level loss is then defined as a cross-entropy alignment between the sharpened teacher distribution and student predictions:
\begin{equation}
\mathcal{L}_{\text{patch}}
=
- \mathbb{E} \left[ 
\sum_{k} 
\tilde{t}_{\text{patch}, k}
\log 
\left( 
\text{softmax}(\tilde{s}_{\text{patch}})_k 
\right)
\right].
\end{equation}
This objective encourages the student to match the teacher's local structural representation of the 3D MRI, complementing the global semantic alignment. The student is optimised with three objectives: CLS-token self-distillation ($L_{\mathrm{global}}$), patch-token self-distillation ($L_{\mathrm{patch}}$), and masked voxel reconstruction ($L_{\mathrm{rec}}$), combined as: 
\begin{equation}
    \mathcal{L} = \mathcal{L}_{\text{global}} + \mathcal{L}_{\text{patch}} + \lambda_{\text{rec}}\mathcal{L}_{\text{rec}}.
\end{equation}
Where $\lambda_{\mathrm{rec}}$ is a configurable hyperparameter (set to 100 in all experiments) that upscales the reconstruction term to a comparable magnitude to the distillation losses, compensating for the smaller per-voxel MSE values that arise from averaging over large 3D volumes. $\lambda_{\mathrm{rec}} = 100$ was selected to bring the reconstruction loss to a comparable magnitude to the distillation losses, following established practice for loss-scale matching in multi-objective SSL.

\noindent \textbf{Knowledge Distillation:} We adopted an online self-distillation framework, where both teacher and student networks were trained jointly using the same 2,002 pretraining MRIs and shared the same architecture. The teacher weights were updated online using an exponential moving average (EMA) of the student parameters. 

\noindent \textbf{Implementation:} Pretraining was performed with a  single 40 GB A100 GPU and we used a batch size of 1 alongside gradient accumulation over 8 iterations. This accounted for the high memory requirements of 3D MRI volumes, enabling an effective batch size of 8, whilst remaining within GPU memory constraints. A gradient accumulation of 8 was selected as a good balance between training speed and increased effective batch size. This strategy also helped stabilise optimisation compared to using a purely batch size 1 update regime. However, gradient accumulation is not exactly equivalent to a true batch size of 8 as gradients are accumulated over sequential mini-batches, while batch-dependent operations, such as teacher centering and EMA updates, are still computed on each mini-batch separately. Training stability was further supported through AdamW optimisation with a learning rate of $5\times10^{-4}$, learning rate warm-up (following a cosine scheduler after 10 warm up epochs), cosine decay scheduling, and EMA teacher updates with gradually increasing momentum ($0.996$ to $1.0$). 

\subsection{Fine-Tuning for Downstream Tasks}

We evaluate \textit{VolTA-3D} on three representative tasks:  (i) sex classification on IXI,  (ii) healthy vs.\ Alzheimer’s disease (AD) classification on AIBL, and (iii) hippocampus segmentation on the HarP-labelled ADNI subset.  
In all experiments, the pretrained VolTA-3D encoder is used as the feature backbone, with weights either frozen or partially unfrozen depending on the task.

\noindent\textbf{Classification:} Classification is performed using a 3D ViT-based classifier. We compare three model initialisation strategies: (1) \textit{VolTA-3D}: ViT encoder initialised with pretrained VolTA-3D weights, (2) \textit{ViT-Scratch}: identical architecture initialised from scratch, and (3) \textit{SwinTransformer-Scratch}: Swin transformer encoder \cite{hatamizadeh2022swinunetrswintransformers} initialised from scratch.

All models are trained for 50~epochs on MRI volumes of size $96 \times 96 \times 96$ using 5-fold cross-validation. Training uses the AdamW optimiser with a learning rate of $1\times10^{-4}$ and a multi-class cross-entropy loss (class-balanced weighted for AIBL). Model performance is assessed using the uncalibrated area under the receiver operating curve (AUROC), averaged across all folds.

\noindent\textbf{Reduced-Data Evaluation:} To assess robustness in low-data regimes, we additionally evaluate sex classification using only 20\%, 40\%, 60\%, 80\%, and 100\% of the available training data. AUROC, averaged across all folds,  is reported for each subset.

\noindent\textbf{Segmentation:} For hippocampus segmentation, we adopt MONAI’s 3D UNETR, using a ViT encoder (convolutional patch embedding, instance normalisation) and a CNN-based decoder trained with Dice + cross-entropy loss. We evaluate three variants: (1) \textit{VolTA-3D}: UNETR \cite{hatamizadeh2022unetr} with encoder initialised from pretrained VolTA-3D weights, (2) \textit{UNETR-Scratch}: identical UNETR architecture with random initialisation, and (3) \textit{3D U-Net-Scratch} ~\cite{34} : CNN with random initialisation and encoder channels [16, 32, 64, 128, 256] and strides [2, 2, 2, 2].

During training, we crop $160 \times 160 \times 160$ volumes to $96 \times 96 \times 96$. The segmentation head predicts three channels (background, left hippocampus, right hippocampus), with background excluded from loss and evaluation. Validation uses a 20\% hold-out split, and full-volume predictions use sliding-window inference. Models train for 500~epochs.

\section{Results}
\subsection{Classification}
\begin{table}[htbp]
\caption{Classification performance on IXI sex classification. 
All results are reported as mean $\pm$ SD across 5-fold cross-validation. A default decision threshold of 0.5 was used. Sensitivity is the female detection rate and specificity is the male detection rate.}
\begin{center}
\begin{tabular}{|l|c|c|c|}
\hline
\textbf{Model} & \textbf{AUROC} & \textbf{Sensitivity} & \textbf{Specificity} \\
\hline

\textbf{VolTA-3D (proposed)} 
& $\mathbf{0.98 \pm 0.01}$ 
& $\mathbf{0.92 \pm 0.06}$ 
& $0.80 \pm 0.20$\\
\hline

ViT-Scratch 
& $0.94 \pm 0.02$& $0.87 \pm 0.08$& $\mathbf{0.82 \pm 0.14}$\\
\hline

SwinTransformer-Scratch 
& $0.85 \pm 0.03$& $0.52 \pm 0.33$ 
& $0.71 \pm 0.22$ \\
\hline

\end{tabular}
\label{tab:ixi_classification_results}
\end{center}
\end{table}
\begin{table}[htbp]
\caption{Classification performance on AIBL Alzheimer’s disease classification (AD vs Healthy). All results are reported as mean $\pm$ SD across 5-fold cross-validation. Youden’s J threshold optimisation is used due to a heavily imbalanced dataset. Sensitivity is the Alzheimer's detection rate and specificity is the healthy detection rate.}
\label{tab:aibl_ad_results_final}
\begin{center}
\begin{tabular}{|l|c|c|c|}
\hline
\textbf{Model} & \textbf{AUROC} & \textbf{Sensitivity} & \textbf{Specificity} \\
\hline
\textbf{VolTA-3D (proposed)} 
& $\mathbf{0.72 \pm 0.10}$ 
& $\mathbf{0.72 \pm 0.20}$ 
& $0.72 \pm 0.08$\\
\hline
ViT-Scratch 
& $0.67 \pm 0.05$ 
& $0.64 \pm 0.16$ 
& $0.72 \pm 0.14$ \\
\hline
SwinTransformer-Scratch 
& $0.64 \pm 0.03$& $0.58 \pm 0.12$ 
& $\mathbf{0.73 \pm 0.11}$ \\
\hline
\end{tabular}

\end{center}
\end{table}
\begin{figure}[t]
    \centering
    \includegraphics[width=\linewidth]{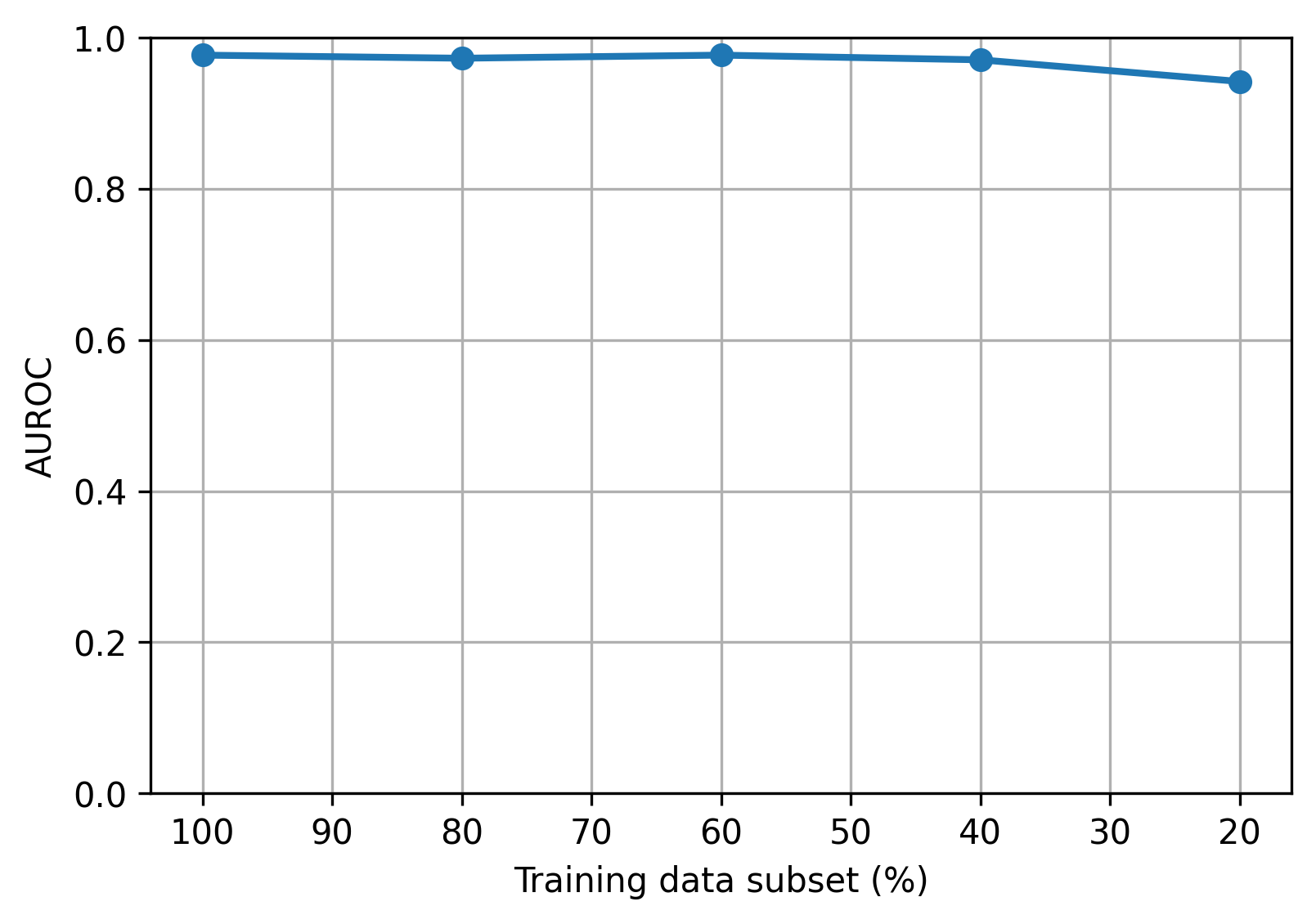}
    \caption{Sex classification performance of VolTA-3D pretrained model on reduced subsets of the IXI dataset.}
    \label{fig:classexpermiment}
\end{figure}

\begin{figure*}[h]
    \centering
    \includegraphics[width=0.95\linewidth]{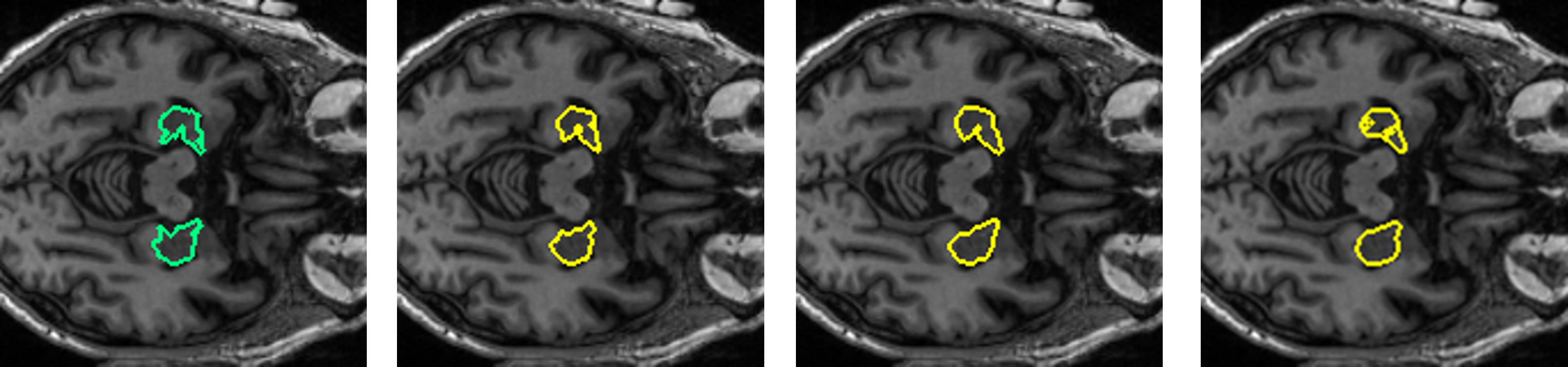}
    \caption{Visualisation of hippocampus segmentation on a consistent MRI slice, using the best-Dice epoch for each model. Left to right: ground truth, VolTA-3D, UNETR-Scratch, 3D U-Net-Scratch.  }
    \label{fig:segmentation}
\end{figure*}

Tables \ref{tab:ixi_classification_results} and \ref{tab:aibl_ad_results_final} report cross-validated AUROC, sensitivity, and specificity for sex and AD classification. For sex classification, VolTA-3D achieves the highest AUROC with lowest variance (0.98 ± 0.01), as well as the highest sensitivity (0.92 ± 0.06), also with this lowest variance. As this is a fairly balanced dataset, the default 0.5 decision threshold was used. These results indicate improved feature discriminability and generalisation. Specificity was improved in the ViT-Scratch model (VolTA-3D: 0.80 ± 0.20, ViT-Scratch: 0.82 ± 0.14), however, the difference was smaller than the comparative improvement in VolTA-3D's AUROC and sensitivity. 
Similarly, for AD classification, VolTA-3D yields the highest AUROC (0.72 ± 0.10), outperforming both comparisons by a larger margin. Although SD is higher, this is offset by the AUROC improvement. VolTA-3D also achieves the highest sensitivity (0.72 ± 0.20), but a slightly lower specificity than the SwinTransformer-Scratch (VolTA-3D: 0.72 ± 0.08, SwinTransformer-Scratch: 0.73 ± 0.11). The difference between the specificty scores across the three models is marginal compared to the difference between the AUROC and sensitivity, in which VolTA-3D shows the strongest performance. 

\subsection{Reduced-Data Evaluation}
Figure \ref{fig:classexpermiment} shows that VolTA-3D, trained on only 20\% of the data, outperforms SwinTransformer-Scratch and matches the performance of ViT-Scratch trained on the full dataset, demonstrating strong suitability to low-data environments. 

\subsection{Segmentation}
Figure \ref{fig:segmentation} visualises model segmentation against the ground truth label, where VolTA-3D shows sharper boundary alignment and less smoothing than the comparative models, particularly relative to the 3D U-Net. Overall, Table \ref{tab:seg_hippocampus_results} summarises hippocampus segmentation performance. Dice was used to select the best epoch. The VolTA-3D pretrained UNETR achieved the highest mean Dice (0.87),  outperforming the randomly initialised UNETR (0.85) and 3D U-Net (0.83).  3D U-Net showed higher boundary precision (but with greater variance), likely due to smoother convolutional upsampling. These results indicate that VolTA-3D pretraining provides stronger transferable anatomical priors than training  ViTs or CNNs from scratch.

\begin{table}[htbp]
\caption{Hippocampus segmentation performance on the HarP-labelled dataset. Reported metrics are mean across subjects with $\pm$ SD per epoch.}
\centering
\begin{tabular}{|l|c|c|}
\hline
\textbf{Model} & \textbf{Dice }& \textbf{HD95 (mm)} \\
\hline
\textbf{VolTA-3D (proposed)} & $\mathbf{0.87 \pm 0.01}$& $2.49 \pm 0.96$\\\hline
{UNETR-Scratch}& $0.85 \pm 0.01$& $\mathbf{3.49 \pm 1.28}$\\\hline
{3D U-Net-Scratch}& $0.83 \pm 0.01$& $1.82 \pm 0.23$\\
\hline
\end{tabular}
\label{tab:seg_hippocampus_results}
\end{table}

\section{Discussion}
VolTA-3D consistently outperformed randomly initialised Vision Transformer, Swin Transformer, and convolutional neural network baselines across both segmentation and classification tasks. These models were deliberately chosen to provide broad architecture comparison, rather than to compete with task-optimised state-of-the-art (SOTA) methods. In this context, VolTA-3D outperforming the Swin Transformer trained from scratch is notably informative, as it contrasts a global-context ViT with a hierarchical transformer architecture employing shifted-window attention~\cite{hatamizadeh2022swinunetrswintransformers}. The observed enhanced performance of VolTA-3D, whilst modest in segmentation, was consistently replicated across cross-validation folds, indicating the potential of this pretraining methodology to enable greater generalisation of brain MRI models to diverse downstream tasks. 

Performance variances across downstream tasks were consistent with known task complexity. Sex classification achieved higher performance than Alzheimer’s disease classification, reflecting the more subtle and heterogeneous morphological differences associated with Alzheimer’s disease~\cite{Ottoy2025}. The lower sensitivity compared to specificity in table \ref{tab:aibl_ad_results_final} is expected for this dataset, as it represents the proportion of true Alzheimer's MRI correctly predicted in a dataset with only 17\% Alzheimer's class. This also indicates the requirement for Youden’s J threshold optimisation. VolTA-3D shows particularly strong performance compared to the comparative models on this more challenging Alzheimer’s disease classification task, suggesting that the benefits of our self-supervised pretraining may be amplified in settings where discriminative features are subtle and difficult to capture through supervised learning alone. 

As shown in table \ref{tab:seg_hippocampus_results}, whilst VolTA-3D demonstrates strong performance considering the Dice score, the 3D U-Net baseline obtained a lower HD95 value. Dice measures the accuracy of the overlap between the ground truth and the prediction, whereas HD95 measures the 95th percentile of distances between the boundaries of the ground truth and the prediction \cite{Sghirripa2025HippocampusSeg}. 

Whilst these are valuable metrics to measure segmentation performance, they require interpretation. HD95 is a boundary-sensitive metric and may still be influenced by local boundary inaccuracies, particularly in small and anatomically intricate structures such as the hippocampus. In a 3D volume, a model can achieve strong overall performance whilst incurring larger errors in a few isolated areas, potentially due to image quality, partial volume effects, or low signal-to-noise ratio. Consequently, relatively small boundary discrepancies may produce comparatively larger changes in HD95, despite good overall segmentation \cite{karimi2019reducinghausdorffdistancemedical}. 

Several limitations are acknowledged. VolTA-3D is not intended to be directly comparable with SOTA models that are optimised for individual downstream tasks, nor does this study aim to establish absolute performance benchmarks. In particular, established pipelines such as FreeSurfer~\cite{freesurfer2012} remain widely used for hippocampal segmentation. The hippocampus was selected due to its extensive study in the literature and the availability of HarP-labelled data ~\cite{boccardi2015harp}. This choice enables a controlled evaluation of transferability, rather than a direct challenge to specialised segmentation systems. That being said, there remains scope for improvement in these systems, as evidenced by the development of the HarP dataset in response to longstanding inconsistencies in hippocampal boundary definitions. This highlights both the inherent difficulty of hippocampal segmentation and the importance of improving the robustness and transferability of learned representations. 

Overall, the objective of this work is to evaluate whether a unified self-supervised 3D pretraining strategy, building upon the original 2D application of Diverse Concept Modelling ~\cite{dicom2024},  can yield representations that generalise across heterogeneous, out-of-distribution tasks. The consistent improvements over randomly initialised baselines across classification and segmentation, spanning multiple model architectures, indicate that VolTA-3D provides a promising foundation for developing more generalisable 3D brain MRI models.

\section{Conclusion}
This work presents VolTA-3D, a self-supervised 3D Vision Transformer pretraining framework that extends Diverse Concept Modelling \cite{dicom2024} to volumetric brain MRI through the integration of Gaussian mixture masked modelling, local and global token alignment, and online self-distillation. 

Across both segmentation and classification tasks, VolTA-3D consistently improved performance over randomly initialised baselines spanning multiple downstream architectures, including Vision Transformers, Swin Transformers, and convolutional neural networks. These findings suggest that unified self-supervised pretraining can learn transferable volumetric representations that generalise across heterogeneous and out-of-distribution brain MRI tasks, without task-specific optimisation.

Although not intended to compete directly with task-specialised state-of-the-art systems, VolTA-3D demonstrates the potential of scalable self-supervised 3D pretraining for developing more robust and generalisable brain MRI models. Future work will investigate larger-scale pretraining, broader downstream evaluation, and architectural refinement to further improve representation quality and transferability in clinically relevant neuroimaging applications. 

\section*{Acknowledgment}

No funding was received for conducting this study. The authors have no relevant financial or non-financial interests to disclose.

\bibliographystyle{IEEEbib}
\bibliography{refs}

\end{document}